\documentclass[11pt]{article}
\pdfoutput=1

\usepackage{acl}

\usepackage{times}
\usepackage{latexsym}

\usepackage[T1]{fontenc}

\usepackage[utf8]{inputenc}

\usepackage{microtype}

\usepackage{tikz}
\usetikzlibrary{shapes.geometric, arrows, calc, backgrounds}
\tikzstyle{filler} = [rectangle, rounded corners, minimum width=0.1cm, minimum height=0.1cm,text centered, opacity=0.0]
\tikzstyle{words} = [rectangle, rounded corners, minimum width=1.5cm, minimum height=0.5cm,text centered, draw=red!80!purple!80, fill=red!30!purple!25]
\tikzstyle{contextual} = [rectangle, minimum width=2cm, minimum height=1cm, text centered, draw=orange!80, fill=orange!30]
\tikzstyle{static} = [rectangle, minimum width=2cm, minimum height=1cm, text centered, draw=green!80!lime!80, fill=green!30!lime!30]
\tikzstyle{operation} = [ellipse, minimum width=1.5cm, minimum height=1cm, text centered, draw=cyan!60!teal!60, fill=cyan!20!teal!20]
\tikzstyle{label} = [circle, text centered, text=black]
\tikzstyle{arrow} = [draw=gray!50!darkgray, thick,->,>=stealth]

\usepackage{arydshln} 

\newcommand{\xlmr}{XLM\mbox{-}R}
\newcommand{\mxtos}{X2S-M}
\newcommand{\maxtos}{X2S-MA}

\title{Combining Static and Contextualised Multilingual Embeddings
}

\author{Katharina Hämmerl$^1$
  \and Jindřich Libovický$^{1,2}$
  \and Alexander Fraser$^1$ \\
$^1$Center for Information and Language Processing, LMU Munich, Germany \\
$^2$Faculty of Mathematics and Physics, Charles University, Czech Republic \\
  \texttt{\{haemmerl,fraser\}@cis.lmu.de} \\
  \texttt{libovicky@ufal.mff.cuni.cz} \\
  }

\begin{document}
\maketitle
\begin{abstract}
Static and contextual multilingual embeddings have complementary strengths.
Static embeddings, while less expressive than contextual language models, can be more straightforwardly aligned across multiple languages.
We combine the strengths of static and contextual models to improve
multilingual representations.
We extract static embeddings for 40 languages from \xlmr, validate those embeddings with cross-lingual word retrieval, and then align them using VecMap.
This results in high-quality, highly multilingual static embeddings.
Then we apply a novel continued pre-training approach to \xlmr, leveraging the high quality alignment of our static embeddings to better align the representation space of \xlmr. We show positive results for multiple complex semantic tasks.
We release the static embeddings and the continued pre-training code.\footnote{\href{https://github.com/KathyHaem/combining-static-contextual}{\texttt{github.com/KathyHaem/combining-static\-contextual}}}
Unlike most
previous work, our continued pre-training approach does not require parallel text.
\end{abstract}

\section{Introduction}

Multilingual contextual encoders like \xlmr\ \citep{conneau-etal-2020-unsupervised} and mBERT \citep{devlin-etal-2019-bert},
despite being trained without parallel data, exhibit ``surprising'' cross-linguality \citep{wu-dredze-2019-beto, conneau-etal-2020-emerging} and have demonstrated strong performance on
multilingual
and cross-lingual tasks (e.g., \citealp{hu2020xtreme, lauscher-etal-2020-zero, kurfali-ostling-2021-probing, turc2021revisiting}).
However, their \textit{language-neutrality}, meaning how well languages are aligned with each other, has clear limits (\citealp{libovicky-etal-2020-language,cao2020multilingual}, inter alia).
In particular, more typologically distant language pairs tend to be less well-aligned than more similar ones, affecting transfer performance.

By contrast, cross-lingual alignment is well-studied for static embeddings (e.g., \citealp{mikolov2013exploiting, artetxe-etal-2018-vecmap-sup, vulic-etal-2020-good}).
They can be aligned using simple transformations,
resulting in high quality multilingual embeddings.
However, static embeddings are considerably less expressive than contextual models and have in many applications been superseded by them.

\begin{figure}
\centering
\begin{tikzpicture}[node distance=1.3cm]
\node (x2sma1) [static] {X2S-MA};
\node (fill1) [filler, left of=x2sma1] {};
\node (vecmap) [operation, left of=fill1] {VecMap};
\node (x2sm) [static] at ($(vecmap) - (0,1.4) $) {X2S-M};
\node (fill2) [filler, right of=x2sm] {};
\node (xlmr1) [contextual, right of=fill2] {\xlmr};
\node (sent) [words, below of=xlmr1] {<s> Die Katze ist gestreift. </s>};
\node (a) [label] at ($(x2sma1) + (2,0) $) {\textbf{a)}};
\draw [arrow] (vecmap) -- (x2sma1);
\draw [arrow] (x2sm) -- (vecmap);
\draw [arrow] (xlmr1) -- (x2sm);
\draw [arrow] (sent) -- (xlmr1);

\node (word1) [words] at ($(sent) - (1.5, 1.5) $) {chat};
\node (x2sma) [static, below of=word1] {X2S-MA};
\node (fill) [filler, right of=x2sma] {};
\node (xlmr) [contextual, right of=fill] {\xlmr};
\node (word2) [words, above of=xlmr] {<s> chat </s>};
\node (alignment) [operation, below of=fill] {CCA/MSE};
\node (b) [label] at ($(word1) - (2,0) $) {\textbf{b)}};
\draw [arrow] (word2) -- (xlmr);
\draw [arrow] (word1) -- (x2sma);
\draw [arrow] (x2sma) -- (alignment);
\draw [arrow] (xlmr) -- (alignment);

 \begin{scope}[on background layer]
    \node (word1b) [words, opacity=0.6] at ($(word1) + (0.1,0.1) $) {};
    \node (word1c) [words, opacity=0.3] at ($(word1) + (0.2,0.2) $) {};
    \node (word2b) [words, opacity=0.6, minimum width=2.35cm] at ($(word2) + (0.1,0.1) $) {};
    \node (word2c) [words, opacity=0.3, minimum width=2.35cm] at ($(word2) + (0.2,0.2) $) {};
    \node (sentb) [words, opacity=0.6, minimum width=5.22cm, minimum height=0.58cm] at ($(sent) + (0.1,0.1) $) {};
    \node (sentc) [words, opacity=0.3, minimum width=5.22cm, minimum height=0.58cm] at ($(sent) + (0.2,0.2) $) {};
    \node (x2smb) [static, opacity=0.7] at ($(x2sm) + (-0.5,0.1) $) {};
    \node (x2smc) [static, opacity=0.4] at ($(x2sm) + (0.4,-0.2) $) {};
    \node (x2smab) [static, opacity=0.7] at ($(x2sma) + (0.1,0.1) $) {};
    \node (x2smac) [static, opacity=0.4] at ($(x2sma) + (0.2,0.2) $) {};
    \node (x2sma1b) [static, opacity=0.7] at ($(x2sma1) + (0.1,0.1) $) {};
    \node (x2sma1c) [static, opacity=0.4] at ($(x2sma1) + (0.2,0.2) $) {};
  \end{scope}

\end{tikzpicture}
\caption{\textbf{a)} We feed sentences from 40 monolingual corpora to \xlmr, extracting partially aligned multilingual static embeddings (\mxtos). Then, we use VecMap to align the embeddings further, giving us \maxtos. The German example sentence reads `\textit{The cat is striped}'.
\textbf{b)} Taking the representations of words from \maxtos\ and \xlmr, we train the contextual model representations to be more similar to the well-aligned static embeddings via an alignment loss (CCA or MSE). The French example `chat' means `\textit{cat}'.
}
\label{fig:diagram}
\end{figure}
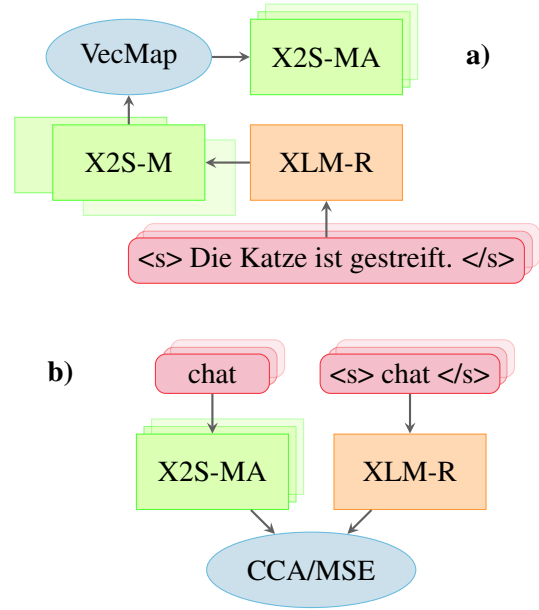

This paper aims to combine the strengths of static and contextual models,
and explore how they may benefit from each other.
Our methods require no parallel corpus.
Figure~\ref{fig:diagram} shows a schematic of our two-part approach with an example:
The words `Katze' in Figure~1a and `chat' in Figure~1b both mean `\textit{cat}'.
While creating \mxtos\ in Figure~1a, static vectors are learned for both words in their respective language embeddings.
We then align those embeddings with VecMap, obtaining \maxtos.
In Fig~1b, we train the contextualised embeddings of `Katze' and `chat' to be closer to their well-aligned \maxtos\ vectors, improving the alignment of the contextualised embeddings.

Monolingual static embeddings have been extracted from BERT by \citet{gupta-jaggi-2021-obtaining}.
We show that their approach can be applied to multilingual embeddings.
We distill static embeddings for 40 languages from \xlmr,
showing that the resulting embeddings are already somewhat cross-lingually aligned, but that their alignment can be improved using established tools (Figure~1a; \S~\ref{sec:extract-embeds}).
These vectors are of high monolingual and cross-lingual quality despite being distilled using only 1M sentences per language.
Second, we present a novel continued pre-training approach for the contextual model, combining masked language modelling (MLM) with an alignment loss that leverages the aligned static embeddings (Figure~1b; \S~\ref{sec:cont-pretraining}).
This results in improved multilingual contextualised embeddings which work
well for complex \mbox{semantic tasks}.

\section{Related Work}
\label{sec:cont-vs-static}

\xlmr\ \citep{conneau-etal-2020-unsupervised} and mBERT \citep{devlin-etal-2019-bert} have been successful in multi- and cross-lingual transfer despite being trained only on monolingual corpora.
However, the 100 languages in \xlmr---or 104 in mBERT---are not represented equally well~(cf.~\citealp{wu-dredze-2020-languages}),
either in terms of data size or downstream performance.
Both \citet{singh-etal-2019-bert} and \citet{libovicky-etal-2020-language} found that mBERT clusters its representations of languages in a way that mirrors typological language family trees.
However, representations being well-aligned across languages is related to better cross-lingual transfer performance.
Therefore,
this property limits the model's transfer ability,
putting target languages which are more distant from the source language
at a disadvantage.

In comparison, static embeddings are far less resource-intensive than contextual models, both at training and inference time.
They can be trained with smaller data and achieve good representation quality where a Transformer model would be under-trained.
Where time, data, or computational resources are limited, this makes static embeddings an attractive approach.
Also,
some NLP tasks rely on static embeddings in their formulation, such as lexical evaluation tasks, approaches comparing vector spaces to detect domain shift \citep{beyer-etal-2020-embedding} or linguistic change \citep{shoemark-etal-2019-room}, or some bias detection and removal tasks (e.g., \citealp{kaneko-bollegala-2019-gender, manzini-etal-2019-black}).
Importantly for us, cross-lingual alignment has been studied extensively in static embeddings (e.g.,~\citealp{artetxe-etal-2018-vecmap-sup, artetxe-etal-2018-robust, joulin-etal-2018-loss}).
Especially those languages that are ill-represented in the massively multilingual model can benefit from using well-aligned static embeddings.
In summary, static and contextual representations have complementary strengths, which we aim to \mbox{combine}.

Recently, cross-lingual alignment objectives have been used to train multilingual contextual models from scratch \citep{hu-etal-2021-explicit, chi-etal-2021-improving}, to align the outputs of monolingual
models \citep{aldarmaki-diab-2019-context, wang-etal-2019-cross}, or to apply a post-hoc alignment to a multilingual
model after pre-training \citep{zhao-etal-2021-inducing, cao2020multilingual, wu-dredze-2020-explicit,kvapilikova-etal-2020-unsupervised, ouyang-etal-2021-ernie, alqahtani-etal-2021-using-optimal}.
These works typically use objectives that rely on translated or induced sentence pairs, such as translation language modelling (TLM; \citealp{lample-conneau-2019-xlm}).
\citet{dou-neubig-2021-word} and \citet{nagata-etal-2020-supervised} focus on word alignment as a task and fine-tune the models on word alignment gold data, though \citet{dou-neubig-2021-word} also use the approach for XNLI.
\citet{gritta-iacobacci-2021-xeroalign} use translated task data to encourage a task-specific alignment of \xlmr.
Some use word-aligned corpora (e.g., \citealp{wang-etal-2019-cross}), while others use parallel sentences plus unsupervised word alignment \citep{alqahtani-etal-2021-using-optimal,chi-etal-2021-improving}. \citet{ouyang-etal-2021-ernie} introduce backtranslation to the alignment process, but still use some parallel data.
\citet{kvapilikova-etal-2020-unsupervised} instead create a synthetic parallel corpus, using this with TLM.

By contrast, we propose an alternate objective that relies on aligned static embedding spaces instead of sentence pairs.
Our alignment approach is a post-hoc tuning of the contextual model using \emph{no parallel corpora} at any point.
This difference allows us to apply the alignment to many more languages than most related work.
For example, \citet{wang-etal-2019-cross} use up to 18 languages, \citet{chi-etal-2021-improving} use 15 with parallel data though 94 in total, \citet{hu-etal-2021-explicit} use 15, while other related work often uses 4--9 languages, with a significant focus on European languages.

\section{Static Embeddings from \xlmr}
\label{sec:extract-embeds}

\citet{gupta-jaggi-2021-obtaining} extracted English static embeddings from BERT and RoBERTa.
They showed that their CBOW-like training scales better with more data and outperforms an aggregation approach to extracting static embeddings~\citep{bommasani-etal-2020-interpreting}.
In their system, X2Static, the context vector from which to predict the target word is given by the average of all vectors in the sentence without the target word.
The method uses ten negative samples per target and calculates the loss based on similarity scores.
However, they only evaluated their method on English.
We extract this type of static embeddings from a multilingual contextual model.

\subsection{Extraction and Alignment Process}

We choose 40 languages for static embeddings extraction (full list in Appendix~\ref{app:langs}).
As the multilingual contextual model, we use \xlmr.
From preliminary experimentation, we determined how best to extract multilingual embeddings from the model:
First, using X2Static \citep{gupta-jaggi-2021-obtaining} worked better than aggregation \citep{bommasani-etal-2020-interpreting} even with a small amount of data.
One important difference with Gupta and Jaggi's work is that for our task the sentence-level variant of X2Static worked better than the paragraph-level version.
Crucially, we also found that embeddings extracted from layer~6 of \xlmr\ performed noticeably better than embeddings extracted from the output layer.
The latter fits with findings for mBERT by \citet{muller-etal-2021-first} that the middle layers are more multilingually aligned.
Due to the large number of languages and having limited data for some of them, we decided to use only up to 1M sentences per language for extraction.
See Appendix \ref{app:prelim-experiments} for more detailed reasoning on these choices.

For the full set of embeddings, we used data from the reconstructed CC100 corpus \citep{wenzek-etal-2020-ccnet}.
We filtered out headlines and too-short sentences heuristically.
See Appendix~\ref{sec:app-tok-seg} for data sampling and processing details.
We refer to the newly extracted embeddings as \textbf{\mxtos}\ for \textbf{X2S}tatic-\textbf{M}ultilingual.

In a second step, we align \mxtos\ using VecMap \citep{artetxe-etal-2018-vecmap-sup} and a set of unsupervised dictionaries that we had previously induced from experiments aligning fasttext vectors~\citep{bojanowski2017enriching} with unsupervised VecMap \citep{artetxe-etal-2018-robust}.
We refer to the aligned embeddings as \textbf{\maxtos}\ (\textbf{X2S}tatic-\textbf{M}ultilingually-\textbf{A}ligned).

\subsection{Embedding Evaluation}
\label{subsec:validate}

\begin{table}
\centering
\begin{tabular}{lcc}
\hline
\textbf{Model} & \textbf{en-xx} & \textbf{xx-en}\\
\hline
fasttext$_{unsup}$ & 54.71 & 58.26 \\
\mxtos & 52.11 & 59.00 \\
\maxtos & 58.41 & 65.60 \\ \hdashline
MUSE~\citep{conneau2018word} & 58.88 & 65.21 \\
RCSLS~\citep{joulin-etal-2018-loss} & \textbf{67.47} & \textbf{71.70} \\
\hline
\end{tabular}
\caption{Results from MUSE BLI tasks. Scores are averaged over those language pairs present in all models. Even before alignment (\mxtos), the embeddings derived from \xlmr\ are competitive with fasttext vectors aligned using unsupervised VecMap (fasttext$_{unsup}$). After alignment and selection (\maxtos), they are on-par with the supervised embeddings released by MUSE despite using much smaller data to train.
We show per-language results in Table~\ref{tab:muse-crossling-perlang}.
}
\label{tab:muse-crossling}
\end{table}

\begin{table}
\centering
\begin{tabular}{lcc}
\hline
\textbf{Model} & \textbf{cross-lingual} & \textbf{monolingual} \\
\hline
fasttext$_{unsup}$ & 0.712 & \textbf{0.743} \\
\mxtos & 0.708 & 0.699 \\
\maxtos & 0.713 & 0.706 \\
\hdashline
MUSE & 0.707 & 0.728 \\
RCSLS & \textbf{0.714} & 0.718 \\
\hline
\end{tabular}
\caption{Average monolingual and cross-lingual scores on SemEval 2017 Task 2
\citep{camacho-collados-etal-2017-semeval}.
See Tables~\ref{tab:semeval20172-cross} and~\ref{tab:semeval20172-mono} for detailed results.
}
\label{tab:semeval20172-summary}
\end{table}

We validate our static embeddings using the MUSE benchmark \citep{conneau2018word}, which includes bilingual dictionary induction (BLI) tasks for 28 of the 40 languages we use, and on SemEval 2017 Task~2 \citep{camacho-collados-etal-2017-semeval}, monolingual and cross-lingual word similarity.
Additionally, we
conduct
a comparative evaluation of the supervised MUSE embeddings and the supervised RCSLS embeddings by \citet{joulin-etal-2018-loss}.

Tables~\ref{tab:muse-crossling} and~\ref{tab:semeval20172-summary} show that after alignment and selection (\maxtos), our vectors perform similarly to the supervised embeddings released by MUSE.
We also contrast \mxtos\ and \maxtos\ against the fasttext embeddings that were used to induce the dictionaries mentioned above.
On the cross-lingual tasks, \maxtos\ performs on par with the fasttext embeddings; on the monolingual tasks, fasttext clearly outperforms \mxtos\ and \maxtos.
Note, however, that SemEval Task~2 only contains data for five of the 40 languages we experiment with.

For most languages, alignment improves BLI by at least a few points, with differences as large as 17 points for Bengali and Hindi (Appendix, Table~\ref{tab:muse-crossling-perlang}).
Such large gaps underline the fact that the alignment of \xlmr\ is suboptimal for these languages.
Notable exceptions are Korean, Thai, Tagalog, and Vietnamese, where the embeddings showed some success before alignment but were not useful afterwards.
It may be that the induced dictionaries did not work well for these languages or that the static embedding spaces were too different
(cf.\ \citealp{vulic-etal-2020-good}).
In these cases, we use the ``unaligned'' embeddings for further experiments.

\section{Cross-Linguality Transfer to \xlmr}
\label{sec:cont-pretraining}

Since our static embeddings are of reasonably high quality after extraction and their cross-linguality can be further improved using established methods, we now ask whether the language neutrality
of the Transformer model can in turn be improved via indirect transfer from our
aligned static embeddings.

\subsection{Continued Pre-Training}

We mix an alignment loss with masked language modelling (MLM).
For the alignment loss, we sample word-vector pairs from our static embeddings, encode the word using
\xlmr, and mean-pool the contextual representations over the subword tokens.
We then compare this representation to the sampled static vector using one of two loss terms:

\paragraph{1) MSE.} We use mean squared error (MSE), i.e., an element-wise comparison of the static and contextual representations.
This works only if the static vector dimension matches the model's hidden size.
\paragraph{2) DCCA.} The second option is a correlation loss (deep canonical correlation analysis; \citealp{andrew2013deepCCA}; implementation from \citealp{arjmand2020dgcca-pytorch}).
Standard CCA \citep{Hotelling1936RelationsBT} takes two continuous representations of related data and linearly transforms them to create two maximally correlated views.
In deep CCA, the linear transformations are replaced by deep networks, which can be optimised on mini-batches.
In our case, we treat the contextual model as one of the two deep models, and replace the other with the static embeddings.
We back-propagate the loss only to the deep model.

We train with two sets of static vectors: Fasttext aligned with unsupervised VecMap (fasttext$_{unsup}$), and our aligned and selected \maxtos\ vectors.
The former have 300 dimensions and so can only be used with DCCA; the latter have 768 dimensions and can thus be used with either loss.

Additionally, we use MLM during training to ensure that the model retains its contextual capabilities.
See Appendix~\ref{app:training-details} for training details.
As a second baseline, we also continue the pre-training with only MLM on our selected languages for the same number of update steps.
This ensures that any improvements from our proposed model are not merely a result of
carrying out further MLM training
in these languages.

\subsection{Downstream Tasks}

\begin{table*}
\centering
\begin{tabular}{lccccccc}
\hline
\textbf{Model} & \textbf{XQuAD} & \textbf{TyDiQA} & \textbf{PAN-X} & \textbf{UD-POS} & \textbf{Tatoeba} & \textbf{avg} \\
\hline
\xlmr & 70.51 & 48.91 & 60.40 & 72.92 & 50.35 & 60.62 \\
+MLM & 70.50 & 48.15 & 61.80 & \textbf{72.97} & 60.87 & 62.86 \\
+fasttext$_{DCCA}$ & 70.84 & \textbf{52.47} & 61.84 & 72.09 & 59.99 & 63.45 \\
+\maxtos$_{MSE}$ & 70.42 & 49.20 & 62.62 & 72.95 & 10.05 & 53.05 \\
+\maxtos$_{DCCA}$ & \textbf{70.92} & 51.02 & \textbf{62.73} & 72.09 & \textbf{68.06} & \textbf{64.96} \\
\hline
\end{tabular}
\caption{Downstream evaluation results.
For the QA and sequence tagging tasks, we report F1 scores averaged over three fine-tuning runs. For Tatoeba we report accuracy. +fasttext$_{DCCA}$ means continued pre-training was done using MLM and DCCA with the aligned fasttext vectors, and analogously for +\maxtos$_{MSE}$ and +\maxtos$_{DCCA}$.
See appendix Tables~\ref{tab:xquad-perlang}-\ref{tab:tatoeba-crossling-perlang} for per-language results.}
\label{tab:downstream}
\end{table*}

For our downstream evaluation tasks, we follow the fine-tuning procedures shown in the repository for~\citet{hu2020xtreme} for better comparability.
We use a zero-shot transfer setting, i.e., we fine-tune only on English data but evaluate on all test sets.
We report mean F1 score over all test sets and three fine-tuning runs for all tasks except Tatoeba, which uses accuracy as its metric and no fine-tuning.

\paragraph{Question Answering.} We use two extractive QA tasks, XQuAD \citep{artetxe-etal-2020-cross} and TyDiQA-GoldP \citep{clark-etal-2020-tydi}.
For XQuAD, the SQuAD v1.1 \citep{rajpurkar-etal-2016-squad} training set is used.
TyDiQA includes its own training set.

\paragraph{Sequence Labelling.} We experiment with the PAN-X \citep{pan-etal-2017-cross} named entity recognition and the UD-POS part-of-speech tagging tasks. The annotated data for UD-POS are taken from Universal Dependencies v2.5 \citep{zeman-etal-2019-ud25}.

\paragraph{Tatoeba}\hspace{-1em} is a sentence retrieval task compiled by \citet{artetxe-schwenk-2019-massively}.
It does not need fine-tuning, instead using the cosine similarity of the mean-pooled layer 7 hidden states for retrieval.

\subsection{Results and Discussion}

Table~\ref{tab:downstream} shows our downstream task results along with the average over all evaluated tasks.
As expected, our second baseline with additional MLM in the affected languages can improve slightly over the unmodified \xlmr.
However, our proposed training with a DCCA loss improves further over both baselines, except on UD-POS.
This shows that the improvement is not merely a result of specialisation on the task languages, but that our alignment loss improves the model's language-neutrality.

Although the fasttext$_{unsup}$ vectors performed very well in Section~\ref{subsec:validate}, using them in continued pre-training is less effective than using \maxtos.
\maxtos\ has the advantage of having the same dimension as the model hidden size, as well as being derived from \xlmr\ itself, both of which likely make it easier to transfer their alignment signal to the contextual model.

While both Tatoeba and the QA tasks favour DCCA, PAN-X improves regardless of the alignment loss used with \maxtos, and UD-POS performance even degrades when using DCCA.
We speculate that this is caused by the different task types requiring different strengths of the model.
Further, UD-POS is a syntactic task, and the strength of the static embeddings is semantic.

The sentence retrieval task, presumably because it relies directly on the cosine similarity of hidden representations, is highly sensitive to changes in the representation.
If only a few dimensions change significantly, this could vastly improve---or ``break''---alignment according to cosine similarity.
By contrast, the tasks using fine-tuning are more stable.
It may also be that although the continued pre-training with DCCA improves the alignment of \xlmr, fine-tuning for tasks on English data then primarily changes the English representation space again, leading to forgetting.
This prompts the question whether the model could in future benefit from using the alignment loss alongside fine-tuning.
Additionally, the static embeddings may be improved further by training them on more data per language, leading to an even better signal for \xlmr.
Recent work also shows that some outlier dimensions in contextual models can obscure representational quality, suggesting that ``accounting for rogue dimensions'' \citep[p.4527]{timkey-van-schijndel-2021-bark} when learning static embeddings may help as well.

\section{Conclusions}

We have extracted high-quality, highly multilingual static embeddings from \xlmr\ using a modified version of X2Static and only 1M sentences of data per language.
Our vectors have reasonable cross-lingual quality immediately after extraction, but we are able to improve their performance using alignment with dictionaries induced from fasttext vectors using VecMap.
No parallel corpus was needed for this process.
Our final models perform competitively with supervised vectors from MUSE, and outperform both MUSE and RCSLS---or provide models at all---for a number of lower- and medium-resource languages.

Further, we proposed a continued pre-training approach that pairs a novel alignment loss with MLM.
Using the DCCA loss, we can improve the language-neutrality of \xlmr, benefitting downstream performance on semantic tasks.

\section*{Ethical Considerations}

Much NLP research is highly English-centric, with a small number of other high-resource languages also benefitting, and the vast majority of languages being left behind or excluded \citep{joshi-etal-2020-state}.
This applies to the multilingual contextual model that we extend, in that high-resource languages are also overrepresented in its training data, and most languages are not part of the model at all.
As well, in the zero-shot transfer tasks we evaluate on, the ``source language'' is English.
Similarly, the BLI datasets we use are mostly xx-en language pairs.
Although this paper makes an effort to reduce the gap between higher- and lower-resource languages, we remain part of this paradigm.
We would like to more strongly focus on low-resource languages in future work.

\section*{Acknowledgements}

The work was supported by the European Research Council (ERC) under the
European Union’s Horizon 2020 research and innovation programme (No.~640550) and by the German Research Foundation (DFG; grant FR
2829/4-1).

\bibliography{anthology,custom}

\begin{thebibliography}{62}
\expandafter\ifx\csname natexlab\endcsname\relax\def\natexlab#1{#1}\fi

\bibitem[{Aldarmaki and Diab(2019)}]{aldarmaki-diab-2019-context}
Hanan Aldarmaki and Mona Diab. 2019.
\newblock \href {https://doi.org/10.18653/v1/N19-1391} {Context-aware
  cross-lingual mapping}.
\newblock In \emph{Proceedings of the 2019 Conference of the North {A}merican
  Chapter of the Association for Computational Linguistics: Human Language
  Technologies, Volume 1 (Long and Short Papers)}, pages 3906--3911,
  Minneapolis, Minnesota. Association for Computational Linguistics.

\bibitem[{Alqahtani et~al.(2021)Alqahtani, Lalwani, Zhang, Romeo, and
  Mansour}]{alqahtani-etal-2021-using-optimal}
Sawsan Alqahtani, Garima Lalwani, Yi~Zhang, Salvatore Romeo, and Saab Mansour.
  2021.
\newblock \href {https://doi.org/10.18653/v1/2021.findings-emnlp.329} {Using
  optimal transport as alignment objective for fine-tuning multilingual
  contextualized embeddings}.
\newblock In \emph{Findings of the Association for Computational Linguistics:
  EMNLP 2021}, pages 3904--3919, Punta Cana, Dominican Republic. Association
  for Computational Linguistics.

\bibitem[{Andrew et~al.(2013)Andrew, Arora, Bilmes, and
  Livescu}]{andrew2013deepCCA}
Galen Andrew, Raman Arora, Jeff Bilmes, and Karen Livescu. 2013.
\newblock Deep canonical correlation analysis.
\newblock In \emph{Proceedings of Machine Learning Research}, volume 28 (3),
  pages 1247--1255, Atlanta, Georgia, USA. PMLR.

\bibitem[{Arjmand(2020)}]{arjmand2020dgcca-pytorch}
Armin Arjmand. 2020.
\newblock \href {https://github.com/arminarj/DeepGCCA-pytorch} {Dgcca-pytorch}.

\bibitem[{Artetxe et~al.(2018{\natexlab{a}})Artetxe, Labaka, and
  Agirre}]{artetxe-etal-2018-vecmap-sup}
Mikel Artetxe, Gorka Labaka, and Eneko Agirre. 2018{\natexlab{a}}.
\newblock \href
  {https://www.aaai.org/ocs/index.php/AAAI/AAAI18/paper/view/16935/16781}
  {Generalizing and improving bilingual word embedding mappings with a
  multi-step framework of linear transformations}.

\bibitem[{Artetxe et~al.(2018{\natexlab{b}})Artetxe, Labaka, and
  Agirre}]{artetxe-etal-2018-robust}
Mikel Artetxe, Gorka Labaka, and Eneko Agirre. 2018{\natexlab{b}}.
\newblock \href {https://doi.org/10.18653/v1/P18-1073} {A robust self-learning
  method for fully unsupervised cross-lingual mappings of word embeddings}.
\newblock In \emph{Proceedings of the 56th Annual Meeting of the Association
  for Computational Linguistics (Volume 1: Long Papers)}, pages 789--798,
  Melbourne, Australia. Association for Computational Linguistics.

\bibitem[{Artetxe et~al.(2020)Artetxe, Ruder, and
  Yogatama}]{artetxe-etal-2020-cross}
Mikel Artetxe, Sebastian Ruder, and Dani Yogatama. 2020.
\newblock \href {https://doi.org/10.18653/v1/2020.acl-main.421} {On the
  cross-lingual transferability of monolingual representations}.
\newblock In \emph{Proceedings of the 58th Annual Meeting of the Association
  for Computational Linguistics}, pages 4623--4637, Online. Association for
  Computational Linguistics.

\bibitem[{Artetxe and Schwenk(2019)}]{artetxe-schwenk-2019-massively}
Mikel Artetxe and Holger Schwenk. 2019.
\newblock \href {https://doi.org/10.1162/tacl_a_00288} {Massively multilingual
  sentence embeddings for zero-shot cross-lingual transfer and beyond}.
\newblock \emph{Transactions of the Association for Computational Linguistics},
  7:597--610.

\bibitem[{Beyer et~al.(2020)Beyer, Kauermann, and
  Sch{\"u}tze}]{beyer-etal-2020-embedding}
Anne Beyer, G{\"o}ran Kauermann, and Hinrich Sch{\"u}tze. 2020.
\newblock \href {https://aclanthology.org/2020.lrec-1.296} {Embedding space
  correlation as a measure of domain similarity}.
\newblock In \emph{Proceedings of the 12th Language Resources and Evaluation
  Conference}, pages 2431--2439, Marseille, France. European Language Resources
  Association.

\bibitem[{Bojanowski et~al.(2017)Bojanowski, Grave, Joulin, and
  Mikolov}]{bojanowski2017enriching}
Piotr Bojanowski, Edouard Grave, Armand Joulin, and Tomas Mikolov. 2017.
\newblock \href {http://arxiv.org/abs/1607.04606} {Enriching word vectors with
  subword information}.

\bibitem[{Bommasani et~al.(2020)Bommasani, Davis, and
  Cardie}]{bommasani-etal-2020-interpreting}
Rishi Bommasani, Kelly Davis, and Claire Cardie. 2020.
\newblock \href {https://doi.org/10.18653/v1/2020.acl-main.431} {{I}nterpreting
  {P}retrained {C}ontextualized {R}epresentations via {R}eductions to {S}tatic
  {E}mbeddings}.
\newblock In \emph{Proceedings of the 58th Annual Meeting of the Association
  for Computational Linguistics}, pages 4758--4781, Online. Association for
  Computational Linguistics.

\bibitem[{Camacho-Collados et~al.(2017)Camacho-Collados, Pilehvar, Collier, and
  Navigli}]{camacho-collados-etal-2017-semeval}
Jose Camacho-Collados, Mohammad~Taher Pilehvar, Nigel Collier, and Roberto
  Navigli. 2017.
\newblock \href {https://doi.org/10.18653/v1/S17-2002} {{S}em{E}val-2017 task
  2: Multilingual and cross-lingual semantic word similarity}.
\newblock In \emph{Proceedings of the 11th International Workshop on Semantic
  Evaluation ({S}em{E}val-2017)}, pages 15--26, Vancouver, Canada. Association
  for Computational Linguistics.

\bibitem[{Cao et~al.(2020)Cao, Kitaev, and Klein}]{cao2020multilingual}
Steven Cao, Nikita Kitaev, and Dan Klein. 2020.
\newblock \href {http://arxiv.org/abs/2002.03518} {Multilingual alignment of
  contextual word representations}.

\bibitem[{Chi et~al.(2021)Chi, Dong, Zheng, Huang, Mao, Huang, and
  Wei}]{chi-etal-2021-improving}
Zewen Chi, Li~Dong, Bo~Zheng, Shaohan Huang, Xian-Ling Mao, Heyan Huang, and
  Furu Wei. 2021.
\newblock \href {https://doi.org/10.18653/v1/2021.acl-long.265} {Improving
  pretrained cross-lingual language models via self-labeled word alignment}.
\newblock In \emph{Proceedings of the 59th Annual Meeting of the Association
  for Computational Linguistics and the 11th International Joint Conference on
  Natural Language Processing (Volume 1: Long Papers)}, pages 3418--3430,
  Online. Association for Computational Linguistics.

\bibitem[{Clark et~al.(2020)Clark, Choi, Collins, Garrette, Kwiatkowski,
  Nikolaev, and Palomaki}]{clark-etal-2020-tydi}
Jonathan~H. Clark, Eunsol Choi, Michael Collins, Dan Garrette, Tom Kwiatkowski,
  Vitaly Nikolaev, and Jennimaria Palomaki. 2020.
\newblock \href {https://doi.org/10.1162/tacl_a_00317} {{T}y{D}i {QA}: A
  benchmark for information-seeking question answering in typologically diverse
  languages}.
\newblock \emph{Transactions of the Association for Computational Linguistics},
  8:454--470.

\bibitem[{Conneau et~al.(2020{\natexlab{a}})Conneau, Khandelwal, Goyal,
  Chaudhary, Wenzek, Guzm{\'a}n, Grave, Ott, Zettlemoyer, and
  Stoyanov}]{conneau-etal-2020-unsupervised}
Alexis Conneau, Kartikay Khandelwal, Naman Goyal, Vishrav Chaudhary, Guillaume
  Wenzek, Francisco Guzm{\'a}n, Edouard Grave, Myle Ott, Luke Zettlemoyer, and
  Veselin Stoyanov. 2020{\natexlab{a}}.
\newblock \href {https://doi.org/10.18653/v1/2020.acl-main.747} {Unsupervised
  cross-lingual representation learning at scale}.
\newblock In \emph{Proceedings of the 58th Annual Meeting of the Association
  for Computational Linguistics}, pages 8440--8451, Online. Association for
  Computational Linguistics.

\bibitem[{Conneau et~al.(2018)Conneau, Lample, Ranzato, Denoyer, and
  Jégou}]{conneau2018word}
Alexis Conneau, Guillaume Lample, Marc'Aurelio Ranzato, Ludovic Denoyer, and
  Hervé Jégou. 2018.
\newblock \href {http://arxiv.org/abs/1710.04087} {Word translation without
  parallel data}.

\bibitem[{Conneau et~al.(2020{\natexlab{b}})Conneau, Wu, Li, Zettlemoyer, and
  Stoyanov}]{conneau-etal-2020-emerging}
Alexis Conneau, Shijie Wu, Haoran Li, Luke Zettlemoyer, and Veselin Stoyanov.
  2020{\natexlab{b}}.
\newblock \href {https://doi.org/10.18653/v1/2020.acl-main.536} {Emerging
  cross-lingual structure in pretrained language models}.
\newblock In \emph{Proceedings of the 58th Annual Meeting of the Association
  for Computational Linguistics}, pages 6022--6034, Online. Association for
  Computational Linguistics.

\bibitem[{Devlin et~al.(2019)Devlin, Chang, Lee, and
  Toutanova}]{devlin-etal-2019-bert}
Jacob Devlin, Ming-Wei Chang, Kenton Lee, and Kristina Toutanova. 2019.
\newblock \href {https://doi.org/10.18653/v1/N19-1423} {{BERT}: Pre-training of
  deep bidirectional transformers for language understanding}.
\newblock In \emph{Proceedings of the 2019 Conference of the North {A}merican
  Chapter of the Association for Computational Linguistics: Human Language
  Technologies, Volume 1 (Long and Short Papers)}, pages 4171--4186,
  Minneapolis, Minnesota. Association for Computational Linguistics.

\bibitem[{Dou and Neubig(2021)}]{dou-neubig-2021-word}
Zi-Yi Dou and Graham Neubig. 2021.
\newblock \href {https://aclanthology.org/2021.eacl-main.181} {Word alignment
  by fine-tuning embeddings on parallel corpora}.
\newblock In \emph{Proceedings of the 16th Conference of the European Chapter
  of the Association for Computational Linguistics: Main Volume}, pages
  2112--2128, Online. Association for Computational Linguistics.

\bibitem[{Eberhard et~al.(2021)Eberhard, Simons, and Fennig}]{ethnologue}
David~M. Eberhard, Gary~F. Simons, and Charles~D. Fennig, editors. 2021.
\newblock \href {http://www.ethnologue.com} {\emph{Ethnologue: Languages of the
  World. Twenty-fourth edition.}}
\newblock SIL International. Online version.

\bibitem[{Gritta and Iacobacci(2021)}]{gritta-iacobacci-2021-xeroalign}
Milan Gritta and Ignacio Iacobacci. 2021.
\newblock \href {https://doi.org/10.18653/v1/2021.findings-acl.32}
  {{X}ero{A}lign: Zero-shot cross-lingual transformer alignment}.
\newblock In \emph{Findings of the Association for Computational Linguistics:
  ACL-IJCNLP 2021}, pages 371--381, Online. Association for Computational
  Linguistics.

\bibitem[{Gupta and Jaggi(2021)}]{gupta-jaggi-2021-obtaining}
Prakhar Gupta and Martin Jaggi. 2021.
\newblock \href {https://doi.org/10.18653/v1/2021.acl-long.408} {Obtaining
  better static word embeddings using contextual embedding models}.
\newblock In \emph{Proceedings of the 59th Annual Meeting of the Association
  for Computational Linguistics and the 11th International Joint Conference on
  Natural Language Processing (Volume 1: Long Papers)}, pages 5241--5253,
  Online. Association for Computational Linguistics.

\bibitem[{Hotelling(1936)}]{Hotelling1936RelationsBT}
Harold Hotelling. 1936.
\newblock Relations between two sets of variates.
\newblock \emph{Biometrika}, 28:321--377.

\bibitem[{Hu et~al.(2021)Hu, Johnson, Firat, Siddhant, and
  Neubig}]{hu-etal-2021-explicit}
Junjie Hu, Melvin Johnson, Orhan Firat, Aditya Siddhant, and Graham Neubig.
  2021.
\newblock \href {https://doi.org/10.18653/v1/2021.naacl-main.284} {Explicit
  alignment objectives for multilingual bidirectional encoders}.
\newblock In \emph{Proceedings of the 2021 Conference of the North American
  Chapter of the Association for Computational Linguistics: Human Language
  Technologies}, pages 3633--3643, Online. Association for Computational
  Linguistics.

\bibitem[{Hu et~al.(2020)Hu, Ruder, Siddhant, Neubig, Firat, and
  Johnson}]{hu2020xtreme}
Junjie Hu, Sebastian Ruder, Aditya Siddhant, Graham Neubig, Orhan Firat, and
  Melvin Johnson. 2020.
\newblock \href {http://arxiv.org/abs/2003.11080} {Xtreme: A massively
  multilingual multi-task benchmark for evaluating cross-lingual
  generalization}.
\newblock \emph{CoRR}, abs/2003.11080.

\bibitem[{Joshi et~al.(2020)Joshi, Santy, Budhiraja, Bali, and
  Choudhury}]{joshi-etal-2020-state}
Pratik Joshi, Sebastin Santy, Amar Budhiraja, Kalika Bali, and Monojit
  Choudhury. 2020.
\newblock \href {https://doi.org/10.18653/v1/2020.acl-main.560} {The state and
  fate of linguistic diversity and inclusion in the {NLP} world}.
\newblock In \emph{Proceedings of the 58th Annual Meeting of the Association
  for Computational Linguistics}, pages 6282--6293, Online. Association for
  Computational Linguistics.

\bibitem[{Joulin et~al.(2018)Joulin, Bojanowski, Mikolov, J{\'e}gou, and
  Grave}]{joulin-etal-2018-loss}
Armand Joulin, Piotr Bojanowski, Tomas Mikolov, Herv{\'e} J{\'e}gou, and
  Edouard Grave. 2018.
\newblock \href {https://doi.org/10.18653/v1/D18-1330} {Loss in translation:
  Learning bilingual word mapping with a retrieval criterion}.
\newblock In \emph{Proceedings of the 2018 Conference on Empirical Methods in
  Natural Language Processing}, pages 2979--2984, Brussels, Belgium.
  Association for Computational Linguistics.

\bibitem[{Junyi(2013)}]{jieba}
Sun Junyi. 2013.
\newblock \href {https://github.com/fxsjy/jieba} {jieba}.

\bibitem[{Kaneko and Bollegala(2019)}]{kaneko-bollegala-2019-gender}
Masahiro Kaneko and Danushka Bollegala. 2019.
\newblock \href {https://doi.org/10.18653/v1/P19-1160} {Gender-preserving
  debiasing for pre-trained word embeddings}.
\newblock In \emph{Proceedings of the 57th Annual Meeting of the Association
  for Computational Linguistics}, pages 1641--1650, Florence, Italy.
  Association for Computational Linguistics.

\bibitem[{Kurfal{\i} and {\"O}stling(2021)}]{kurfali-ostling-2021-probing}
Murathan Kurfal{\i} and Robert {\"O}stling. 2021.
\newblock \href {https://doi.org/10.18653/v1/2021.repl4nlp-1.2} {Probing
  multilingual language models for discourse}.
\newblock In \emph{Proceedings of the 6th Workshop on Representation Learning
  for NLP (RepL4NLP-2021)}, pages 8--19, Online. Association for Computational
  Linguistics.

\bibitem[{Kvapil{\'\i}kov{\'a} et~al.(2020)Kvapil{\'\i}kov{\'a}, Artetxe,
  Labaka, Agirre, and Bojar}]{kvapilikova-etal-2020-unsupervised}
Ivana Kvapil{\'\i}kov{\'a}, Mikel Artetxe, Gorka Labaka, Eneko Agirre, and
  Ond{\v{r}}ej Bojar. 2020.
\newblock \href {https://doi.org/10.18653/v1/2020.acl-srw.34} {Unsupervised
  multilingual sentence embeddings for parallel corpus mining}.
\newblock In \emph{Proceedings of the 58th Annual Meeting of the Association
  for Computational Linguistics: Student Research Workshop}, pages 255--262,
  Online. Association for Computational Linguistics.

\bibitem[{Lample and Conneau(2019)}]{lample-conneau-2019-xlm}
Guillaume Lample and Alexis Conneau. 2019.
\newblock \href {http://arxiv.org/abs/1901.07291} {Cross-lingual language model
  pretraining}.
\newblock \emph{CoRR}, abs/1901.07291.

\bibitem[{Lauscher et~al.(2020)Lauscher, Ravishankar, Vuli{\'c}, and
  Glava{\v{s}}}]{lauscher-etal-2020-zero}
Anne Lauscher, Vinit Ravishankar, Ivan Vuli{\'c}, and Goran Glava{\v{s}}. 2020.
\newblock \href {https://doi.org/10.18653/v1/2020.emnlp-main.363} {From zero to
  hero: {O}n the limitations of zero-shot language transfer with multilingual
  {T}ransformers}.
\newblock In \emph{Proceedings of the 2020 Conference on Empirical Methods in
  Natural Language Processing (EMNLP)}, pages 4483--4499, Online. Association
  for Computational Linguistics.

\bibitem[{Libovick{\'y} et~al.(2020)Libovick{\'y}, Rosa, and
  Fraser}]{libovicky-etal-2020-language}
Jind{\v{r}}ich Libovick{\'y}, Rudolf Rosa, and Alexander Fraser. 2020.
\newblock \href {https://doi.org/10.18653/v1/2020.findings-emnlp.150} {On the
  language neutrality of pre-trained multilingual representations}.
\newblock In \emph{Findings of the Association for Computational Linguistics:
  EMNLP 2020}, pages 1663--1674, Online. Association for Computational
  Linguistics.

\bibitem[{Manzini et~al.(2019)Manzini, Yao~Chong, Black, and
  Tsvetkov}]{manzini-etal-2019-black}
Thomas Manzini, Lim Yao~Chong, Alan~W Black, and Yulia Tsvetkov. 2019.
\newblock \href {https://doi.org/10.18653/v1/N19-1062} {Black is to criminal as
  caucasian is to police: Detecting and removing multiclass bias in word
  embeddings}.
\newblock In \emph{Proceedings of the 2019 Conference of the North {A}merican
  Chapter of the Association for Computational Linguistics: Human Language
  Technologies, Volume 1 (Long and Short Papers)}, pages 615--621, Minneapolis,
  Minnesota. Association for Computational Linguistics.

\bibitem[{McCann(2020)}]{mccann-2020-fugashi}
Paul McCann. 2020.
\newblock \href {https://doi.org/10.18653/v1/2020.nlposs-1.7} {fugashi, a tool
  for tokenizing {J}apanese in python}.
\newblock In \emph{Proceedings of Second Workshop for NLP Open Source Software
  (NLP-OSS)}, pages 44--51, Online. Association for Computational Linguistics.

\bibitem[{Mikolov et~al.(2013)Mikolov, Le, and
  Sutskever}]{mikolov2013exploiting}
Tomas Mikolov, Quoc~V. Le, and Ilya Sutskever. 2013.
\newblock \href {http://arxiv.org/abs/1309.4168} {Exploiting similarities among
  languages for machine translation}.

\bibitem[{Muller et~al.(2021)Muller, Elazar, Sagot, and
  Seddah}]{muller-etal-2021-first}
Benjamin Muller, Yanai Elazar, Beno{\^\i}t Sagot, and Djam{\'e} Seddah. 2021.
\newblock \href {https://aclanthology.org/2021.eacl-main.189} {First align,
  then predict: Understanding the cross-lingual ability of multilingual
  {BERT}}.
\newblock In \emph{Proceedings of the 16th Conference of the European Chapter
  of the Association for Computational Linguistics: Main Volume}, pages
  2214--2231, Online. Association for Computational Linguistics.

\bibitem[{Nagata et~al.(2020)Nagata, Chousa, and
  Nishino}]{nagata-etal-2020-supervised}
Masaaki Nagata, Katsuki Chousa, and Masaaki Nishino. 2020.
\newblock \href {https://doi.org/10.18653/v1/2020.emnlp-main.41} {A supervised
  word alignment method based on cross-language span prediction using
  multilingual {BERT}}.
\newblock In \emph{Proceedings of the 2020 Conference on Empirical Methods in
  Natural Language Processing (EMNLP)}, pages 555--565, Online. Association for
  Computational Linguistics.

\bibitem[{Ouyang et~al.(2021)Ouyang, Wang, Pang, Sun, Tian, Wu, and
  Wang}]{ouyang-etal-2021-ernie}
Xuan Ouyang, Shuohuan Wang, Chao Pang, Yu~Sun, Hao Tian, Hua Wu, and Haifeng
  Wang. 2021.
\newblock \href {https://doi.org/10.18653/v1/2021.emnlp-main.3} {{ERNIE}-{M}:
  Enhanced multilingual representation by aligning cross-lingual semantics with
  monolingual corpora}.
\newblock In \emph{Proceedings of the 2021 Conference on Empirical Methods in
  Natural Language Processing}, pages 27--38, Online and Punta Cana, Dominican
  Republic. Association for Computational Linguistics.

\bibitem[{Pan et~al.(2017)Pan, Zhang, May, Nothman, Knight, and
  Ji}]{pan-etal-2017-cross}
Xiaoman Pan, Boliang Zhang, Jonathan May, Joel Nothman, Kevin Knight, and Heng
  Ji. 2017.
\newblock \href {https://doi.org/10.18653/v1/P17-1178} {Cross-lingual name
  tagging and linking for 282 languages}.
\newblock In \emph{Proceedings of the 55th Annual Meeting of the Association
  for Computational Linguistics (Volume 1: Long Papers)}, pages 1946--1958,
  Vancouver, Canada. Association for Computational Linguistics.

\bibitem[{Paszke et~al.(2019)Paszke, Gross, Massa, Lerer, Bradbury, Chanan,
  Killeen, Lin, Gimelshein, Antiga, Desmaison, Kopf, Yang, DeVito, Raison,
  Tejani, Chilamkurthy, Steiner, Fang, Bai, and Chintala}]{PyTorch}
Adam Paszke, Sam Gross, Francisco Massa, Adam Lerer, James Bradbury, Gregory
  Chanan, Trevor Killeen, Zeming Lin, Natalia Gimelshein, Luca Antiga, Alban
  Desmaison, Andreas Kopf, Edward Yang, Zachary DeVito, Martin Raison, Alykhan
  Tejani, Sasank Chilamkurthy, Benoit Steiner, Lu~Fang, Junjie Bai, and Soumith
  Chintala. 2019.
\newblock Pytorch: An imperative style, high-performance deep learning library.
\newblock In H.~Wallach, H.~Larochelle, A.~Beygelzimer, F.~d\textquotesingle
  Alch\'{e}-Buc, E.~Fox, and R.~Garnett, editors, \emph{Advances in Neural
  Information Processing Systems 32}, pages 8024--8035. Curran Associates, Inc.

\bibitem[{Rajpurkar et~al.(2016)Rajpurkar, Zhang, Lopyrev, and
  Liang}]{rajpurkar-etal-2016-squad}
Pranav Rajpurkar, Jian Zhang, Konstantin Lopyrev, and Percy Liang. 2016.
\newblock \href {https://doi.org/10.18653/v1/D16-1264} {{SQ}u{AD}: 100,000+
  questions for machine comprehension of text}.
\newblock In \emph{Proceedings of the 2016 Conference on Empirical Methods in
  Natural Language Processing}, pages 2383--2392, Austin, Texas. Association
  for Computational Linguistics.

\bibitem[{{\v R}eh{\r u}{\v r}ek and Sojka(2010)}]{rehurek2010gensim}
Radim {\v R}eh{\r u}{\v r}ek and Petr Sojka. 2010.
\newblock {Software Framework for Topic Modelling with Large Corpora}.
\newblock In \emph{{Proceedings of the LREC 2010 Workshop on New Challenges for
  NLP Frameworks}}, pages 45--50, Valletta, Malta. ELRA.

\bibitem[{Rosa(2018)}]{rosa2018wikitext}
Rudolf Rosa. 2018.
\newblock \href {http://hdl.handle.net/11234/1-2735} {Plaintext wikipedia dump
  2018}.
\newblock {LINDAT}/{CLARIAH}-{CZ} digital library at the Institute of Formal
  and Applied Linguistics ({{\'U}FAL}), Faculty of Mathematics and Physics,
  Charles University.

\bibitem[{Rui(2020)}]{icutokenizer}
Ming Rui. 2020.
\newblock \href {https://github.com/mingruimingrui/ICU-tokenizer}
  {Icu-tokenizer}.

\bibitem[{Shoemark et~al.(2019)Shoemark, Liza, Nguyen, Hale, and
  McGillivray}]{shoemark-etal-2019-room}
Philippa Shoemark, Farhana~Ferdousi Liza, Dong Nguyen, Scott Hale, and Barbara
  McGillivray. 2019.
\newblock \href {https://doi.org/10.18653/v1/D19-1007} {Room to {G}lo: A
  systematic comparison of semantic change detection approaches with word
  embeddings}.
\newblock In \emph{Proceedings of the 2019 Conference on Empirical Methods in
  Natural Language Processing and the 9th International Joint Conference on
  Natural Language Processing (EMNLP-IJCNLP)}, pages 66--76, Hong Kong, China.
  Association for Computational Linguistics.

\bibitem[{Singh et~al.(2019)Singh, McCann, Socher, and
  Xiong}]{singh-etal-2019-bert}
Jasdeep Singh, Bryan McCann, Richard Socher, and Caiming Xiong. 2019.
\newblock \href {https://doi.org/10.18653/v1/D19-6106} {{BERT} is not an
  interlingua and the bias of tokenization}.
\newblock In \emph{Proceedings of the 2nd Workshop on Deep Learning Approaches
  for Low-Resource NLP (DeepLo 2019)}, pages 47--55, Hong Kong, China.
  Association for Computational Linguistics.

\bibitem[{Straka and Strakov{\'a}(2017)}]{straka-strakova-2017-tokenizing}
Milan Straka and Jana Strakov{\'a}. 2017.
\newblock \href {https://doi.org/10.18653/v1/K17-3009} {Tokenizing, {POS}
  tagging, lemmatizing and parsing {UD} 2.0 with {UDP}ipe}.
\newblock In \emph{Proceedings of the {C}o{NLL} 2017 Shared Task: Multilingual
  Parsing from Raw Text to Universal Dependencies}, pages 88--99, Vancouver,
  Canada. Association for Computational Linguistics.

\bibitem[{{Text Analysis and Knowledge Engineering
  Lab}(2021)}]{takelab-Spacy-Udpipe}
{Text Analysis and Knowledge Engineering Lab}. 2021.
\newblock \href {https://github.com/TakeLab/spacy-udpipe} {spacy-udpipe}.

\bibitem[{Timkey and van Schijndel(2021)}]{timkey-van-schijndel-2021-bark}
William Timkey and Marten van Schijndel. 2021.
\newblock \href {https://aclanthology.org/2021.emnlp-main.372} {All bark and no
  bite: Rogue dimensions in transformer language models obscure
  representational quality}.
\newblock In \emph{Proceedings of the 2021 Conference on Empirical Methods in
  Natural Language Processing}, pages 4527--4546, Online and Punta Cana,
  Dominican Republic. Association for Computational Linguistics.

\bibitem[{Turc et~al.(2021)Turc, Lee, Eisenstein, Chang, and
  Toutanova}]{turc2021revisiting}
Iulia Turc, Kenton Lee, Jacob Eisenstein, Ming-Wei Chang, and Kristina
  Toutanova. 2021.
\newblock \href {http://arxiv.org/abs/2106.16171} {Revisiting the primacy of
  english in zero-shot cross-lingual transfer}.

\bibitem[{Vuli{\'c} et~al.(2020)Vuli{\'c}, Ruder, and
  S{\o}gaard}]{vulic-etal-2020-good}
Ivan Vuli{\'c}, Sebastian Ruder, and Anders S{\o}gaard. 2020.
\newblock \href {https://doi.org/10.18653/v1/2020.emnlp-main.257} {Are all good
  word vector spaces isomorphic?}
\newblock In \emph{Proceedings of the 2020 Conference on Empirical Methods in
  Natural Language Processing (EMNLP)}, pages 3178--3192, Online. Association
  for Computational Linguistics.

\bibitem[{Wang et~al.(2019)Wang, Che, Guo, Liu, and Liu}]{wang-etal-2019-cross}
Yuxuan Wang, Wanxiang Che, Jiang Guo, Yijia Liu, and Ting Liu. 2019.
\newblock \href {https://doi.org/10.18653/v1/D19-1575} {Cross-lingual {BERT}
  transformation for zero-shot dependency parsing}.
\newblock In \emph{Proceedings of the 2019 Conference on Empirical Methods in
  Natural Language Processing and the 9th International Joint Conference on
  Natural Language Processing (EMNLP-IJCNLP)}, pages 5721--5727, Hong Kong,
  China. Association for Computational Linguistics.

\bibitem[{Wenzek et~al.(2020)Wenzek, Lachaux, Conneau, Chaudhary, Guzm{\'a}n,
  Joulin, and Grave}]{wenzek-etal-2020-ccnet}
Guillaume Wenzek, Marie-Anne Lachaux, Alexis Conneau, Vishrav Chaudhary,
  Francisco Guzm{\'a}n, Armand Joulin, and Edouard Grave. 2020.
\newblock \href {https://aclanthology.org/2020.lrec-1.494} {{CCN}et: Extracting
  high quality monolingual datasets from web crawl data}.
\newblock In \emph{Proceedings of the 12th Language Resources and Evaluation
  Conference}, pages 4003--4012, Marseille, France. European Language Resources
  Association.

\bibitem[{Wolf et~al.(2020)Wolf, Debut, Sanh, Chaumond, Delangue, Moi, Cistac,
  Rault, Louf, Funtowicz, Davison, Shleifer, von Platen, Ma, Jernite, Plu, Xu,
  Le~Scao, Gugger, Drame, Lhoest, and Rush}]{wolf-etal-2020-transformers}
Thomas Wolf, Lysandre Debut, Victor Sanh, Julien Chaumond, Clement Delangue,
  Anthony Moi, Pierric Cistac, Tim Rault, Remi Louf, Morgan Funtowicz, Joe
  Davison, Sam Shleifer, Patrick von Platen, Clara Ma, Yacine Jernite, Julien
  Plu, Canwen Xu, Teven Le~Scao, Sylvain Gugger, Mariama Drame, Quentin Lhoest,
  and Alexander Rush. 2020.
\newblock \href {https://doi.org/10.18653/v1/2020.emnlp-demos.6} {Transformers:
  State-of-the-art natural language processing}.
\newblock In \emph{Proceedings of the 2020 Conference on Empirical Methods in
  Natural Language Processing: System Demonstrations}, pages 38--45, Online.
  Association for Computational Linguistics.

\bibitem[{Wu and Dredze(2019)}]{wu-dredze-2019-beto}
Shijie Wu and Mark Dredze. 2019.
\newblock \href {https://doi.org/10.18653/v1/D19-1077} {Beto, bentz, becas: The
  surprising cross-lingual effectiveness of {BERT}}.
\newblock In \emph{Proceedings of the 2019 Conference on Empirical Methods in
  Natural Language Processing and the 9th International Joint Conference on
  Natural Language Processing (EMNLP-IJCNLP)}, pages 833--844, Hong Kong,
  China. Association for Computational Linguistics.

\bibitem[{Wu and Dredze(2020{\natexlab{a}})}]{wu-dredze-2020-languages}
Shijie Wu and Mark Dredze. 2020{\natexlab{a}}.
\newblock \href {https://doi.org/10.18653/v1/2020.repl4nlp-1.16} {Are all
  languages created equal in multilingual {BERT}?}
\newblock In \emph{Proceedings of the 5th Workshop on Representation Learning
  for NLP}, pages 120--130, Online. Association for Computational Linguistics.

\bibitem[{Wu and Dredze(2020{\natexlab{b}})}]{wu-dredze-2020-explicit}
Shijie Wu and Mark Dredze. 2020{\natexlab{b}}.
\newblock \href {https://doi.org/10.18653/v1/2020.emnlp-main.362} {Do explicit
  alignments robustly improve multilingual encoders?}
\newblock In \emph{Proceedings of the 2020 Conference on Empirical Methods in
  Natural Language Processing (EMNLP)}, pages 4471--4482, Online. Association
  for Computational Linguistics.

\bibitem[{Zeman et~al.(2019)Zeman, Nivre et~al.}]{zeman-etal-2019-ud25}
Daniel Zeman, Joakim Nivre, et~al. 2019.
\newblock \href {http://hdl.handle.net/11234/1-3105} {Universal dependencies
  2.5}.
\newblock {LINDAT}/{CLARIAH}-{CZ} digital library at the Institute of Formal
  and Applied Linguistics ({{\'U}FAL}), Faculty of Mathematics and Physics,
  Charles University.

\bibitem[{Zhao et~al.(2021)Zhao, Eger, Bjerva, and
  Augenstein}]{zhao-etal-2021-inducing}
Wei Zhao, Steffen Eger, Johannes Bjerva, and Isabelle Augenstein. 2021.
\newblock \href {https://doi.org/10.18653/v1/2021.starsem-1.22} {Inducing
  language-agnostic multilingual representations}.
\newblock In \emph{Proceedings of *SEM 2021: The Tenth Joint Conference on
  Lexical and Computational Semantics}, pages 229--240, Online. Association for
  Computational Linguistics.

\end{thebibliography}

\appendix

\section{List of Languages}
\label{app:langs}

We list all languages used in our experiments in Table~\ref{tab:langs}.

\begin{table}[th!]
\centering
\begin{tabular}{lll}
\hline
\textbf{Language} & \textbf{Code} & \textbf{Family}\\
\hline
Afrikaans & af & IE: Germanic \\
Arabic & ar & Semitic \\
Bulgarian & bg & IE: Slavic \\
Bengali & bn & IE: Indo-Aryan \\
German & de & IE: Germanic \\
Greek & el & IE: Greek \\
English & en & IE: Germanic \\
Spanish & es & IE: Romance \\
Estonian & et & Uralic \\
Basque & eu & Isolate \\
Farsi & fa & IE: Iranian \\
Finnish & fi & Uralic \\
French & fr & IE: Romance \\
Hebrew & he & Semitic \\
Hindi & hi & IE: Indo-Aryan \\
Hungarian & hu & Uralic \\
Indonesian & id & Malayo-Polynesian \\
Italian & it & IE: Romance \\
Japanese & ja & Japonic \\
Javanese & jv & Malayo-Polynesian \\
Georgian & ka & Kartvelian \\
Kazakh & kk & Turkic \\
Korean & ko & Koreanic \\
Malayalam & ml & Dravidian \\
Marathi & mr & IE: Indo-Aryan \\
Malay & ms & Malayo-Polynesian \\
Burmese & my & Sino-Tibetan \\
Dutch & nl & IE: Germanic \\
Portuguese & pt & IE: Romance \\
Russian & ru & IE: Slavic \\
Swahili & sw & Niger-Congo \\
Tamil & ta & Dravidian \\
Telugu & te & Dravidian \\
Thai & th & Kra-Dai \\
Tagalog & tl & Malayo-Polynesian \\
Turkish & tr & Turkic \\
Urdu & ur & IE: Indo-Aryan \\
Vietnamese & vi & Mon-Khmer \\
Yoruba & yo & Niger-Congo \\
Mandarin & zh & Sino-Tibetan \\
\hline
\end{tabular}
\caption{List of languages used with their ISO codes and language families \citep{ethnologue}. IE stands for Indo-European.
}
\label{tab:langs}
\end{table}

\section{Design Choices for Embedding Extraction}
\label{app:prelim-experiments}

\paragraph{Preliminary Experiments.}
We arrived at many of our design choices through preliminary experiments on English, French and German.
Of course, these are highly related languages; however, they allowed for easier debugging in the early stages of embedding extraction.
For these experiments, we used 100k paragraphs per language taken from the Wikipedia dataset by \citet{rosa2018wikitext}, and applied the data filtering methods proposed by the respective authors.
We first tested the approach by \citet{bommasani-etal-2020-interpreting} on all layers of \xlmr\ and found the best BLI
performances in layer six.
However, we also found that the method from \citet{gupta-jaggi-2021-obtaining} tended to outperform the pooling approach even on this small data size while scaling better according to the authors.
Inspired by the results on the pooling method, we decided to test the second approach on layer six as well, leading to better BLI results.
Rather than expend the (GPU) time to train embeddings on every layer, we then experimented with different alignment algorithms before deciding on VecMap for its slight performance advantage and quick training time.

\paragraph{Data Size.}
We decided to use no more than 1M sentences per language partly to upper-bound resource consumption (note that this still amounts to 40M sentences of training data), and partly to put high- and low-resource languages on a somewhat more even footing.
For example, \citet{vulic-etal-2020-good} suggest that vastly different training data sizes make embedding alignment more difficult.
They also find that at least the BLI performance of high-frequency words starts to saturate when the aligned embeddings were trained from scratch using around 1M sentences.
Since our embeddings additionally have something of a head start due to initialisation from \xlmr, 1M sentences per language would seem to be a reasonable data size.

\paragraph{Dimensionality.}
The high dimensionality of the vectors is on one hand a direct result of the extraction method, but on the other hand we believe it may be an advantage for our subsequent alignment experiments, since having the same dimensionality as the contextual model seems to increase the stability of our continued pre-training.
Quite likely the high dimensionality is a disadvantage for the BLI performance of these vectors due to hubness issues; however, their performance is remarkably competitive considering this.

\section{Data Sampling and Processing Details for \mxtos}
\label{sec:app-tok-seg}

\paragraph{Data Sampling.}
After sampling data from the reconstructed CC100 corpus \citep{wenzek-etal-2020-ccnet}, we do sentence segmentation and tokenisation (see the list of languages and tools below),
then filter the data heuristically:
Like~\citet{bommasani-etal-2020-interpreting}, we discard sentences with fewer than seven tokens.
We also keep only sentences from paragraphs with at least two sentences, avoiding, for example, headlines.

\paragraph{Segmentation and Tokenisation Tools.}
af, ar, bg, de, en, el, es, et, eu, fa, fi, fr, he, hi, hu, id, it, ko, mr, nl, pt, ru, ta, te, tr, ur, vi: Spacy-UDPipe~\citep{straka-strakova-2017-tokenizing, takelab-Spacy-Udpipe} version 1.0.0 for both sentence segmentation and tokenisation. ja: ICU-tokenizer~\citep{icutokenizer} version 0.0.1 for sentence segmentation, fugashi~\citep{mccann-2020-fugashi} version 1.1.1 for tokenisation. zh: ICU-tokenizer for sentence segmentation, jieba~\citep{jieba} version 0.42.1 for tokenisation. bn, jv, ka, kk, ml, ms, my, sw, th, tl, yo: ICU-tokenizer for both.

\section{Continued Pre-Training Details}
\label{app:training-details}

We start from \xlmr$_\textsc{Base}$, which has 270M parameters.
At each training step, we mix samples from a text dataset with samples from our static embeddings, computing both a language modelling and an alignment loss.
We use an effective batch size of 64 for MLM and 1024 for the alignment loss.
We use Gensim \citep{rehurek2010gensim} version 4.0.0 to load the static embeddings.
The data for MLM is sampled from concatenated Wikipedia data of all 40 languages.
For this corpus, 100k paragraphs per language were taken from \citet{rosa2018wikitext}.
Each model is trained for 7500 update steps, corresponding to roughly four epochs over our set of static embeddings.
We use the default hyper\-parameters for language modelling in Huggingface Transformers \citep{wolf-etal-2020-transformers} version 4.8.2.
The final checkpoints are selected based on the MLM loss over a separate validation set.
Training was done using PyTorch \citep{PyTorch} in version 1.9.
Each training run was done on a single Nvidia GeForce GTX 1080 Ti GPU.


\begin{table*}
\centering\small
\begin{tabular}{lcccccccccccc}
\hline
\textbf{Model} & \textbf{af-en} &\textbf{en-af} & \textbf{ar-en} & \textbf{en-ar} & \textbf{bg-en} & \textbf{en-bg} & \textbf{bn-en} & \textbf{en-bn} & \textbf{de-en} & \textbf{en-de} & \textbf{el-en} & \textbf{en-el}  \\
\hline
fasttext$_{unsup}$ & 38.49 & 30.38 & 49.02 & 39.07 & 59.99 & 46. 27 & 33.31 & 24.50 & 70.11 & 76.65 & 60.15 & 49.86 \\
\mxtos & 59.94 & 57.01 & 34.88 & 25.59 & 57.10 & 44.72 & 18.03 & 18.04 & 63.34 & 65.69 & 48.22 & 35.95 \\
\maxtos & \textbf{64.37} & \textbf{57.01} & 49.89 & 38.46 & 64.75 & 51.23 & \textbf{36.16} & 33.07 & 69.81 & 73.21 & 58.76 & 47.21 \\
\hdashline
MUSE & -- & -- & 49.87 & 39.74 & 57.53 & 47.27 & -- & -- & 72.67 & 74.67 & 58.47 & 46.27 \\
RCSLS & 40.00 & 36.27 & \textbf{59.57} & \textbf{56.33} & \textbf{65.20} & \textbf{58.20} & 28.41 & \textbf{35.93} & \textbf{77.53} & \textbf{79.20} & \textbf{64.53} & \textbf{55.07} \\
\hline
\end{tabular}

\begin{tabular}{lcccccccccccc}
\hline
\textbf{Model} & \textbf{es-en} & \textbf{en-es} & \textbf{et-en} & \textbf{en-et} & \textbf{fa-en} & \textbf{en-fa} & \textbf{fi-en} & \textbf{en-fi} & \textbf{fr-en} & \textbf{en-fr} & \textbf{he-en} & \textbf{en-he} \\
\hline
fasttext$_{unsup}$ & 77.53 & 79.87 & 49.16 & 38.15 & 38.24 & 35.41 & 51.51 & 44.97 & 77.20 & 80.59 & 54.82 & 44.82 \\
\mxtos & 75.90 & 72.25 & 51.23 & 38.41 & 33.78 & 30.56 & 53.32 & 45.11 & 72.88 & 71.47 & 39.64 & 32.29 \\
\maxtos & 78.88 & 77.11 & \textbf{59.07} & 46.69 & 42.88 & 39.13 & 60.13 & 47.91 & 77.19 & 77.53 & 56.04 & 43.71 \\
\hdashline
MUSE & 83.47 & 81.87 & 45.67 & 37.87 & -- & -- & 59.47 & 48.07 & 82.40 & 82.93 & 54.14 & 44.07 \\
RCSLS & \textbf{87.13} & \textbf{83.73} & 53.67 & \textbf{52.93} & \textbf{44.27} & \textbf{45.33} & \textbf{69.93} & \textbf{61.80} & \textbf{84.73} & \textbf{84.13} & \textbf{59.88} & \textbf{58.53} \\
\hline
\end{tabular}

\begin{tabular}{lcccccccccccc}
\hline
\textbf{Model} & \textbf{hi-en} & \textbf{en-hi} & \textbf{hu-en} & \textbf{en-hu} & \textbf{id-en} & \textbf{en-id} & \textbf{it-en} & \textbf{en-it} & \textbf{ja-en} & \textbf{en-ja} & \textbf{ko-en} & \textbf{en-ko} \\
\hline
fasttext$_{unsup}$ & 48.00 & 38.58 & 58.89 & 54.45 & 63.95 & 66.35 & 72.86 & 78.80 & 40.06 & 45.40 & 0.07 & 0.00 \\
\mxtos & 32.22 & 33.24 & 59.29 & 49.00 & 69.26 & 66.38 & 72.25 & 68.20 & 26.89 & 36.24 & 30.77 & 22.63 \\
\maxtos & \textbf{51.66} & \textbf{48.22} & 65.16 & 55.16 & \textbf{75.22} & 72.36 & 78.25 & 74.80 & \textbf{39.46} & \textbf{45.59} & 27.52 & 24.14 \\
\hdashline
MUSE & -- & -- & 64.87 & 53.87 & 67.93 & 67.40 & 77.87 & 78.60 & -- & -- & -- & -- \\
RCSLS & 46.95 & 44.47 & \textbf{73.00} & \textbf{67.00} & 72.87 & \textbf{72.87} & \textbf{82.73} & \textbf{81.07} & -- & -- & \textbf{36.55} & \textbf{57.47} \\
\hline
\end{tabular}

\begin{tabular}{lcccccccccccc}
\hline
\textbf{Model} & \textbf{ms-en} & \textbf{en-ms} & \textbf{nl-en} & \textbf{en-nl} & \textbf{pt-en} & \textbf{en-pt} & \textbf{ru-en} & \textbf{en-ru} & \textbf{ta-en} & \textbf{en-ta} & \textbf{th-en} & \textbf{en-th} \\
\hline
fasttext$_{unsup}$ & 39.68 & 41.95 & 70.49 & 76.22 & 69.79 & 69.41 & 55.84 & 44.08 & 29.31 & 24.87 & 0.00 & 0.00 \\
\mxtos & 56.89 & 55.99 & 69.50 & 69.58 & 76.47 & 75.04 & 53.07 & 38.14 & 17.68 & 16.26 & 29.06 & 29.69 \\
\maxtos & \textbf{66.04} & \textbf{61.24} & 75.13 & 75.04 & 79.72 & 73.71 & 60.97 & 45.96 & \textbf{31.92} & 30.55 & \textbf{27.13} & 30.02 \\
\hdashline
MUSE & -- & -- & 75.33 & 75.53 & 80.27 & 81.27 & 63.67 & 54.07 & -- & -- & -- & -- \\
RCSLS & -- & -- & \textbf{80.47} & \textbf{79.67} & \textbf{84.60} & \textbf{83.13} & \textbf{70.27} & \textbf{60.93} & 22.84 & \textbf{30.67} & 21.07 & \textbf{32.27} \\
\hline
\end{tabular}

\begin{tabular}{lcccccccccccc}
\hline
\textbf{Model} & \textbf{tl-en} & \textbf{en-tl} & \textbf{tr-en} & \textbf{en-tr} & \textbf{vi-en} & \textbf{en-vi} & \textbf{zh-en} & \textbf{en-zh} \\
\hline
fasttext$_{unsup}$ & 0.00 & 0.00 & 49.12 & 40.58 & 0.00 & 0.00 & 43.90 & 23.70 \\
\mxtos & \textbf{53.99} & \textbf{52.84} & 55.58 & 45.26 & 51.02 & 41.76 & 35.04 & 36.26 \\
\maxtos & 53.49 & 52.75 & 56.20 & 47.75 & 50.00 & 43.77 & 44.89 & 44.70 \\
\hdashline
MUSE & -- & -- & 59.17 & 49.93 & 55.80 & 40.60 & -- & -- \\
RCSLS & 23.60 & 31.87 & \textbf{65.78} & \textbf{59.20} & \textbf{66.93} & \textbf{53.13} & \textbf{48.87} & \textbf{52.40} \\
\hline
\end{tabular}

\caption{Cross-lingual MUSE results, per language with English.}
\label{tab:muse-crossling-perlang}
\end{table*}


\begin{table*}
\centering
\begin{tabular}{lccccccccccc}
\hline
\textbf{Model} & \textbf{de-en} & \textbf{de-es} & \textbf{de-fa} & \textbf{de-it} & \textbf{en-es} & \textbf{en-fa} & \textbf{en-it} & \textbf{es-fa} & \textbf{es-it} & \textbf{fa-it} & \textbf{avg} \\
\hline
fasttext$_{unsup}$ & \textbf{0.74} & \textbf{0.75} & \textbf{0.69} & \textbf{0.72} & \textbf{0.73} & 0.69 & 0.71 & 0.70 & \textbf{0.74} & 0.66 & 0.712 \\
\mxtos & 0.71 & 0.73 & 0.66 & 0.70 & 0.72 & 0.69 & 0.72 & \textbf{0.73} & \textbf{0.74} & 0.69 & 0.708 \\
\maxtos & 0.72 & 0.72 & 0.67 & 0.70 & \textbf{0.73} & 0.71 & 0.73 & 0.72 & \textbf{0.74} & 0.69 & 0.713 \\ \hdashline
MUSE & 0.71 & 0.70 & -- & 0.68 & 0.71 & -- & 0.71 & -- & 0.73 & -- & 0.707 \\
RCSLS & \textbf{0.74} & 0.71 & 0.67 & 0.69 & \textbf{0.73} & \textbf{0.73} & \textbf{0.74} & 0.71 & 0.73 & \textbf{0.70} & \textbf{0.714} \\
\hline
\end{tabular}
\caption{Full cross-lingual results from SemEval 2017 Task 2 \citep{camacho-collados-etal-2017-semeval}.}
\label{tab:semeval20172-cross}
\end{table*}

\begin{table*}
\centering
\begin{tabular}{lccccc}
\hline
\textbf{Model} & \textbf{de} & \textbf{en} & \textbf{es} & \textbf{fa} & \textbf{it} \\
\hline
fasttext$_{unsup}$ & \textbf{0.80} & 0.71 & \textbf{0.76} & \textbf{0.72} & \textbf{0.73} \\
\mxtos & 0.73 & 0.70 & 0.73 & 0.65 & 0.68 \\
\maxtos & 0.73 & \textbf{0.72 }& 0.72 & 0.66 & 0.70 \\
\hdashline
MUSE~\citep{conneau2018word} & 0.73 & \textbf{0.72} & 0.74 & -- & 0.72 \\
RCSLS~\citep{joulin-etal-2018-loss} & 0.73 & \textbf{0.72} & 0.74 & 0.66 & \textbf{0.73} \\
\hline
\end{tabular}
\caption{Full monolingual results from SemEval 2017 Task 2 \citep{camacho-collados-etal-2017-semeval}.}
\label{tab:semeval20172-mono}
\end{table*}


\begin{table*}
\centering\small
\begin{tabular}{lccccccccccc}
\hline
\textbf{Model} & \textbf{ar} & \textbf{de} & \textbf{el} & \textbf{en} & \textbf{es} & \textbf{hi} & \textbf{ru} & \textbf{th} & \textbf{tr} & \textbf{vi} & \textbf{zh} \\
\hline
\xlmr & 65.34 & 74.47 & 72.57 & 83.21 & 76.98 & 67.72 & 74.31 & 67.66 & \textbf{68.55} & 73.66 & 51.09 \\
+MLM & 64.93 & 74.73 & 72.52 & 83.66 & 76.75 & 68.00 & 74.30 & 67.76 & 67.86 & 73.35 & 51.68 \\
+fasttext$_{DCCA}$ & 65.50 & 74.77 & \textbf{73.78} & 83.66 & 76.75 & 68.84 & \textbf{75.06} & 67.35 & 68.30 & \textbf{74.18} & 51.00
 \\
+\maxtos$_{MSE}$ & 64.73 & 74.01 & 72.87 & 83.51 & 76.36 & 67.82 & 74.46 & \textbf{67.77} & 68.04 & 73.78 & 51.30 \\
+\maxtos$_{DCCA}$ & \textbf{65.91} & \textbf{74.83} & 73.05 & \textbf{84.07} & \textbf{77.00} & \textbf{69.29} & 74.26 & 66.99 & \textbf{68.55} & 73.98 & \textbf{52.20} \\
\hline
\end{tabular}
\caption{XQuAD results (F1) per language. Averaged over three fine-tuning runs with different random seeds.}
\label{tab:xquad-perlang}
\end{table*}


\begin{table*}
\centering
\begin{tabular}{lccccccccccc}
\hline
\textbf{Model} & \textbf{ar} & \textbf{bn} & \textbf{en} & \textbf{fi} & \textbf{id} & \textbf{ko} & \textbf{ru} & \textbf{sw} & \textbf{te} \\
\hline
\xlmr & 57.43 & 37.20 & 62.74 & 53.87 & 68.04 & 20.67 & 52.25 & 54.16 & 33.80 \\
+MLM & 57.89 & 35.48 & 62.38 & 51.70 & 66.06 & 21.08 & 52.64 & 54.76 & 31.40 \\
+fasttext$_{DCCA}$ & \textbf{60.96} & \textbf{43.20} & \textbf{63.79} & 56.52 & \textbf{70.72} & \textbf{23.58} & \textbf{55.57} & \textbf{55.37} & \textbf{42.56}
 \\
+\maxtos$_{MSE}$ & 57.46 & 37.59 & 61.16 & 52.95 & 66.77 & 21.73 & 51.63 & 53.10 & \textbf{40.43} \\
+\maxtos$_{DCCA}$  & 58.58 & 42.69 & 63.48 & \textbf{56.78} & 69.02 & 23.11 & 54.55 & 54.90 & 36.04 \\
\hline
\end{tabular}
\caption{TyDiQA results (F1) per language. Averaged over three fine-tuning runs with different random seeds.}
\label{tab:tydiqa-perlang}
\end{table*}


\begin{table*}
\centering
\begin{tabular}{lccccccccccc}
\hline
\textbf{Model} & \textbf{af} & \textbf{ar} & \textbf{bg} & \textbf{bn} & \textbf{de} & \textbf{el} & \textbf{en} & \textbf{es} & \textbf{et} & \textbf{eu} \\
\hline
\xlmr & 74.88 & 46.12 & 77.18 & 67.96 & 74.34 & 72.97 & \textbf{82.83} & 74.52 & 70.44 & 57.75 \\
+MLM & 76.48 & 48.25 & 77.51 & 69.89 & 75.00 & 73.88 & 82.75 & 75.90 & 73.17 & 57.21 \\
+fasttext$_{DCCA}$ & \textbf{77.93} & 47.58 & \textbf{78.00} & 67.27 & \textbf{76.23} & 75.34 & 82.82 & \textbf{79.45} & 74.06 & 61.43 \\
+\maxtos$_{MSE}$ & 76.87 & 47.86 & 77.79 & \textbf{70.69} & 75.58 & \textbf{76.34} & 82.72 & 77.87 & 73.96 & \textbf{61.90} \\
+\maxtos$_{DCCA}$  & 77.50 & \textbf{53.03} & 77.98 & 66.16 & 75.81 & 75.30 & 82.73 & 75.76 & \textbf{74.67} & 60.28 \\
\hline
\end{tabular}

\begin{tabular}{lcccccccccc}
\hline
\textbf{Model} & \textbf{fa} & \textbf{fi} & \textbf{fr} & \textbf{he} & \textbf{hi} & \textbf{hu} & \textbf{id} & \textbf{it} & \textbf{ja} & \textbf{jv} \\
\hline
\xlmr & 49.30 & 74.95 & 77.51 & 51.86 & 66.65 & 76.10 & 48.99 & 77.13 & 19.61 & 57.45 \\
+MLM & 47.72 & 75.52 & \textbf{79.17} & 53.63 & \textbf{68.74} & 76.94 & 50.62 & 77.48 & 18.28 & 58.32 \\
+fasttext$_{DCCA}$ & 47.74 & 76.93 & 78.71 & 56.70 & 66.66 & \textbf{77.27} & 49.35 & \textbf{78.56} & 17.48 & 59.14 \\
+\maxtos$_{MSE}$ & \textbf{55.45} & \textbf{76.30} & 78.83 & \textbf{57.81} & 67.76 & 77.22 & 49.92 & 77.98 & \textbf{20.53} & \textbf{63.28} \\
+\maxtos$_{DCCA}$  & 50.56 & 76.20 & 78.88 & 54.91 & 67.86 & 76.83 & \textbf{55.03} & 78.13 & 17.94 & 58.42 \\
\hline
\end{tabular}

\begin{tabular}{lcccccccccc}
\hline
\textbf{Model} & \textbf{ka} & \textbf{kk} & \textbf{ko} & \textbf{ml} & \textbf{mr} & \textbf{ms} & \textbf{my} & \textbf{nl} &\textbf{pt} & \textbf{ru} \\
\hline
\xlmr & 65.60 & 45.45 & 48.07 & 60.50 & 61.31 & 62.54 & 53.09 & 79.45 & 77.67 & 63.42 \\
+MLM & 67.35 & 51.14 & 51.97 & 63.19 & 61.30 & 67.42 & 52.84 & 80.64 & 79.14 & 62.40 \\
+fasttext$_{DCCA}$ & 67.88 & 51.49 & 47.48 & 51.92 & \textbf{63.13} & 57.89 & 46.19 & \textbf{81.25} & 79.48 & 64.41 \\
+\maxtos$_{MSE}$ & \textbf{69.14} & \textbf{51.76} & \textbf{54.13} & \textbf{64.49} & 62.96 & \textbf{67.43} & \textbf{53.53} & 80.82 & 78.90 & \textbf{64.50} \\
+\maxtos$_{DCCA}$  & 66.49 & 50.59 & 52.55 & 59.64 & 60.35 & 66.94 & 51.79 & 81.06 & \textbf{80.45} & 62.77 \\
\hline
\end{tabular}

\begin{tabular}{lcccccccccc}
\hline
\textbf{Model} & \textbf{sw} & \textbf{ta} & \textbf{te} & \textbf{th} & \textbf{tl} & \textbf{tr} & \textbf{ur} & \textbf{vi} & \textbf{yo} & \textbf{zh} \\
\hline
\xlmr & 63.96 & 54.64 & 48.66 & 3.60 & 71.46 & 74.68 & 54.31 & 68.58 & 34.91 & \textbf{25.47} \\
+MLM & 65.27 & 56.12 & 50.77 & 3.34 & 71.39 & 76.49 & 62.23 & 69.88 & 38.05 & 24.51 \\
+fasttext$_{DCCA}$ & \textbf{66.45} & 57.31 & 53.63 & 3.42 & \textbf{71.78} & \textbf{78.59} & 56.52 & \textbf{71.97} & \textbf{53.07} & 21.26 \\
+\maxtos$_{MSE}$ & 66.35 & \textbf{58.47} & 53.66 & 3.22 & 70.49 & 77.09 & 60.26 & 69.90 & 37.00 & 24.33  \\
+\maxtos$_{DCCA}$ & 65.40 & 56.26 & \textbf{54.61} & 2.19 & 67.65 & 77.53 & \textbf{63.47} & 70.53 & 50.23 & 24.40 \\
\hline
\end{tabular}
\caption{PAN-X results (F1) per language. Averaged over three fine-tuning runs with different random seeds.}
\label{tab:panx-perlang}
\end{table*}


\begin{table*}
\centering
\begin{tabular}{lcccccccccc}
\hline
\textbf{Model} & \textbf{af} & \textbf{ar} & \textbf{bg} & \textbf{de} & \textbf{el} & \textbf{en} & \textbf{es} & \textbf{et} & \textbf{eu} \\
\hline
\xlmr & 88.46 & 67.56 & 88.58 & 88.64 & \textbf{87.79} & \textbf{95.85} & 88.04 & 85.63 & \textbf{69.38} \\
+MLM & 88.75 & 68.21 & \textbf{88.85} & 88.57 & 87.37 & 95.71 & 88.51 & 85.88 & 69.05 \\
+fasttext$_{DCCA}$ & \textbf{88.96} & 67.73 & 88.30 & 88.40 & 87.34 & 95.79 & 87.33 & 85.58 & 68.33 \\
+\maxtos$_{MSE}$ & 88.87 & \textbf{68.43} & 88.55 & \textbf{88.72} & 87.45 & 95.77 & \textbf{88.61} & 85.72 & 69.27 \\
+\maxtos$_{DCCA}$ & 88.50 & 67.45 & 88.11 & 88.22 & 87.26 & 95.69 & 87.87 & \textbf{85.99} & 68.34 \\
\hline
\end{tabular}

\begin{tabular}{lccccccccc}
\hline
\textbf{Model} & \textbf{fa} & \textbf{fi} & \textbf{fr} & \textbf{he} & \textbf{hi} & \textbf{hu} & \textbf{id} & \textbf{it} & \textbf{ja} \\
\hline
\xlmr & 70.16 & 85.60 & 86.00 & 66.96 & 67.83 & \textbf{83.14} & 72.64 & 87.41 & \textbf{24.23} \\
+MLM & 70.14 & \textbf{85.75} & 86.50 & \textbf{68.51} & 68.14 & 83.07 & \textbf{72.59} & 88.46 & 23.59 \\
+fasttext$_{DCCA}$ & 68.70 & 85.69 & 86.20 & 66.33 & 65.70 & 82.87 & 72.64 & 87.32 & 13.89 \\
+\maxtos$_{MSE}$ & \textbf{70.46} & 85.61 & \textbf{86.76} & 67.63 & \textbf{69.30} & 82.82 & \textbf{72.59} & \textbf{88.61} & 20.61 \\
+\maxtos$_{DCCA}$  & 68.81 & 85.74 & 86.38 & 66.34 & 66.01 & 82.89 & 72.82 & 87.43 & 14.12 \\
\hline
\end{tabular}

\begin{tabular}{lccccccccc}
\hline
\textbf{Model} & \textbf{kk} & \textbf{ko} & \textbf{mr} & \textbf{nl} &\textbf{pt} & \textbf{ru}  & \textbf{ta} & \textbf{te} & \textbf{th} \\
\hline
\xlmr & 76.74 & 53.06 & 82.95 & 89.42 & 86.21 & \textbf{89.25} & 62.12 & \textbf{84.90} & 42.36 \\
+MLM & 76.54 & 52.88 & 83.21 & \textbf{89.45} & 86.82 & 89.00 & 61.62 & 83.79 & 42.09 \\
+fasttext$_{DCCA}$ & \textbf{78.09} & 52.86 & 82.86 & 89.35 & 85.70 & 89.11 & \textbf{63.00} & 84.21 & 41.54 \\
+\maxtos$_{MSE}$ & 76.55 & \textbf{53.16} & \textbf{84.19} & \textbf{89.45} & \textbf{87.45} & 89.17 & 61.44 & 84.60 & \textbf{42.62} \\
+\maxtos$_{DCCA}$ & 77.78 & 52.93 & 82.66 & 89.37 & 86.07 & 88.89 & 62.21 & 84.49 & 39.63 \\
\hline
\end{tabular}

\begin{tabular}{lcccccc}
\hline
\textbf{Model} & \textbf{tl} & \textbf{tr} & \textbf{ur} & \textbf{vi} & \textbf{yo} & \textbf{zh} \\
\hline
\xlmr & 88.91 & 74.27 & 56.48 & \textbf{58.59} & 25.29 & \textbf{32.08} \\
+MLM & \textbf{89.42} & 74.20 & 56.58 & 58.21 & 24.38 & 32.06 \\
+fasttext$_{DCCA}$ & 88.22 & 74.53 & 56.06 & 57.62 & 23.76 & 25.02 \\
+\maxtos$_{MSE}$ & 89.21 & 74.19 & \textbf{57.45} & 58.15 & \textbf{25.45} & 28.54 \\
+\maxtos$_{DCCA}$ & 87.44 & \textbf{74.58} & 56.79 & 57.68 & 24.55 & 25.80 \\
\hline
\end{tabular}
\caption{UD-POS results (F1) per language. Averaged over three fine-tuning runs with different random seeds.}
\label{tab:udpos-perlang}
\end{table*}

\begin{table*}
\centering
\begin{tabular}{lcccccccccc}
\hline
\textbf{Model} & \textbf{af} & \textbf{ar} & \textbf{bg} & \textbf{bn} & \textbf{de} & \textbf{el} & \textbf{es} & \textbf{et} & \textbf{eu} \\
\hline
\xlmr & 51.60 & 35.80 & 66.90 & 28.70 & 88.40 & 51.60 & 71.00 & 44.20 & 26.10 \\
+MLM & 65.60 & 46.50 & 74.70 & 41.70 & 91.90 & 61.10 & 79.00 & 55.80 & 38.60 \\
+fasttext$_{DCCA}$ & 70.60 & 47.20 & 78.20 & 44.90 & 95.00 & 68.40 & 85.80 & 63.90 & 44.70 \\
+\maxtos$_{MSE}$ & 10.90 & 3.90 & 17.10 & 2.40 & 42.50 & 5.10 & 15.20 & 7.90 & 7.40 \\
+\maxtos$_{DCCA}$  & \textbf{74.10} & \textbf{57.00} & \textbf{82.10} & \textbf{54.90} & \textbf{95.40} & \textbf{72.50} & \textbf{88.60} & \textbf{75.20} & \textbf{52.50} \\
\hline
\end{tabular}

\begin{tabular}{lccccccccc}
\hline
\textbf{Model} & \textbf{fa} & \textbf{fi} & \textbf{fr} & \textbf{he} & \textbf{hi} & \textbf{hu} & \textbf{id} & \textbf{it} & \textbf{ja} \\
\hline
\xlmr & 64.40 & 63.90 & 72.50 & 51.70 & 50.50 & 58.70 & 68.60 & 64.70 & 52.80 \\
+MLM & 73.50 & 74.60 & 77.90 & 65.10 & 69.10 & 69.90 & 81.10 & 73.40 & 64.20 \\
+fasttext$_{DCCA}$ & 74.60 & 78.60 & 82.30 & 65.50 & 61.90 & 73.30 & 82.80 & 78.50 & 67.00 \\
+\maxtos$_{MSE}$ & 10.50 & 12.70 & 22.20 & 10.10 & 9.00 & 13.40 & 14.30 & 11.50 & 10.00 \\
+\maxtos$_{DCCA}$  & \textbf{79.90} & \textbf{84.30} & \textbf{84.30} & \textbf{71.70} & \textbf{70.10} & \textbf{80.20} & \textbf{86.40} & \textbf{82.30} & \textbf{74.00} \\
\hline
\end{tabular}

\begin{tabular}{lccccccccc}
\hline
\textbf{Model} & \textbf{jv} & \textbf{ka} & \textbf{kk} & \textbf{ko} & \textbf{ml} & \textbf{mr} & \textbf{nl} &\textbf{pt} & \textbf{ru} \\
\hline
\xlmr & 15.12 & 37.13 & 33.22 & 50.10 & 54.73 & 38.00 & 76.80 & 76.60 & 69.80 \\
+MLM & 20.00 & 45.98 & 44.17 & 61.00 & \textbf{64.19} & \textbf{50.70} & 84.60 & 84.40 & 78.50 \\
+fasttext$_{DCCA}$ & 16.10 & 30.56 & 53.39 & 40.40 & 14.56 & 35.40 & 87.20 & 88.30 & 83.00 \\
+\maxtos$_{MSE}$ & 5.37 & 4.96 & 6.09 & 10.50 & 4.51 & 5.30 & 17.80 & 19.70 & 12.50 \\
+\maxtos$_{DCCA}$ & \textbf{22.93} & \textbf{63.81} & \textbf{62.26} & \textbf{63.20} & 25.47 & 34.90 & \textbf{89.30} & \textbf{90.40} & \textbf{85.60} \\
\hline
\end{tabular}

\begin{tabular}{lccccccccc}
\hline
\textbf{Model} & \textbf{sw} & \textbf{ta} & \textbf{te} & \textbf{th} & \textbf{tl} & \textbf{tr} & \textbf{ur} & \textbf{vi} & \textbf{zh} \\
\hline
\xlmr & 15.64 & 25.08 & 30.77 & 34.67 & 29.70 & 54.90 & 31.10 & 67.70 & 59.40 \\
+MLM & 23.59 & 36.16 & 37.61 & 51.28 & 39.90 & 65.20 & \textbf{47.40} & 77.50 & 75.60 \\
+fasttext$_{DCCA}$ & 21.54 & 42.35 & 51.28 & 35.58 & 37.80 & 69.30 & 42.60 & 76.20 & 70.80 \\
+\maxtos$_{MSE}$ & 4.10 & 1.95 & 3.42 & 1.64 & 6.80 & 6.80 & 2.50 & 15.60 & 6.10  \\
+\maxtos$_{DCCA}$ & \textbf{23.85} & \textbf{56.35} & \textbf{59.40} & \textbf{68.43} & \textbf{45.10} & \textbf{78.00} & 45.90 & \textbf{84.40} & \textbf{85.20} \\
\hline
\end{tabular}
\caption{Tatoeba results (accuracy) per language.}
\label{tab:tatoeba-crossling-perlang}
\end{table*}

\end{document}